# A Machine Learning Approach to Comment Toxicity Classification

Navoneel Chakrabarty[1]

[1]Jalpaiguri Government Engineering College, Jalpaiguri, West Bengal, India
nc2012@cse.jgec.ac.in

**Abstract.** Now-a-days, derogatory comments are often made by one another, not only in offline environment but also immensely in online environments like social networking websites and online communities. So, an Identification combined with Prevention System in all social networking websites and applications, including all the communities, existing in the digital world is a necessity. In such a system, the Identification Block should identify any negative online behaviour and should signal the Prevention Block to take action accordingly. This study aims to analyse any piece of text and detecting different types of toxicity like obscenity, threats, insults and identity-based hatred. The labelled Wikipedia Comment Dataset prepared by Jigsaw is used for the purpose. A 6-headed Machine Learning tf-idf Model has been made and trained separately, yielding a Mean Validation Accuracy of 98.08% and Absolute Validation Accuracy of 91.61%. Such an Automated System should be deployed for enhancing healthy online conversation.

## 1 Introduction

Over a decade, social networking and social media have been growing in leaps and bounds. Today, people are able to express themselves and their opinions and also discuss among others via these platforms. In such a scenario, it is quite obvious that debates may arise due to differences in opinion. But often these debates take a dirty side and may result in fights over the social media during which offensive language termed as toxic comments may be used from one side. These toxic comments may be threatening, obscene, insulting or identity-based hatred. So, these clearly pose the threat of abuse and harassment online. Consequently, some people stop giving their opinions or give up seeking different opinions which result in unhealthy and unfair discussion. As a result, different platforms and communities find it very difficult to facilitate fair conversation and are often forced to either limit user comments or get dissolved by shutting down user comments completely. The Conversation AI team, a research group founded by Jigsaw and Google have been working on tools and techniques for providing an environment for healthy communication[1]. They have also built publicly available models through the Perspective API on Comment Toxicity[2].

But these models are sometimes prone to errors and does not provide the option to the users for choosing which type of toxicity, they are interested in finding. So, a more stable and versatile intelligent system is required for Toxic Comment Prevention in social communication. This model reads any piece of text (a text message or any comment appearing in social platform that can be toxic or non-toxic) and detects the type of toxicity it contains. The types of toxicity are simply toxic, severely toxic, obscene, threat, insult and identity-based hate. This overcomes the drawback of the model developed using Perspective API, showing all the types of toxicity contained in the comment.

This paper has been structured as follows: Section 2 throws light on the existing works and approaches used by them as Literature Review, Section 3 describes the Proposed Methodology including the dataset used, data visualizations and model construction, Section 4 elucidates the Individual Training of the Pipelines, Section 5 mentions the Implementation Details, Section 6 deals with the Results, depicting the Model Performance. Finally, it is concluded with future scope or improvement in Section 7.

## 2   Literature Review

Many Machine and Deep Learning Approaches have been attempted for detecting types of toxicity in comments.

- Georgakopoulos et al. proposed a Deep Learning Approach involving Convolutional Neural Networks (CNNs) for text analytics in toxicity classification, obtaining a Mean Accuracy of 91.2%[3].
- Khieu et al. applied various Deep Learning approaches involving Long-Short Term Memory Networks (LSTMs) for the task of classifying toxicity in online comments, obtaining a Label Accuracy of 92.7%[4].
- Chu et al. implemented a Convolutional Neural Network (CNN) with character-level embedding for detecting types of toxicity in online comments, obtaining a Mean Accuracy of 94%[5].
- Kohli et al. proposed a Deep Learning Approach involving Recurrent Neural Network (RNN) Long-Short Term Memory with Custom Embeddings for comment toxicity classification, obtaining a Mean Accuracy of 97.78%[6].

## 3   Proposed Methodology

It consists of 4 subsections: Sub Section 3.1 describes the Dataset used, Sub Section 3.2 deals with Data Visualization, Sub Section 3.3 deals with the Text Preprocessing and Sub Section 3.4 illustrates the Pipelines.

### 3.1   The Dataset

The Wikipedia Talk Page Dataset prepared by Jigsaw and now publicly available at Kaggle is used[7]. The Dataset consists of total 159571 instances with comments and corresponding multiple binomial labels: toxic, severe toxic, obscene, threat, insult and identity hate. Sample instances of the dataset are shown below in Fig 1.

| id | comment_text | toxic | severe_toxic | obscene | threat | insult | identity_hate |
|---|---|---|---|---|---|---|---|
| 0044cf18cc2655b3 | What page shoudld there be for important characters that DON'T reoccur? | 0 | 0 | 0 | 0 | 0 | 0 |
| 00472b8e2d38d1ea | Void, Black Doom, Mephiles, etc | 1 | 0 | 1 | 0 | 1 | 1 |

**Fig 1. Sample Instances of the Dataset**

### 3.2  Data Visualization

Visualizations are done in the form of Histograms showing the distribution of comment lengths over the whole corpus of the dataset in Fig 2. The number of bins in the histogram are found using Freedman-Diaconis Formula given in equation (1).

$$h = 2 * IQR * n^{-1/3} \qquad (1)$$

where,
- IQR is the Inter Quartile Range
- n is the Number of Data Points
- h is the Bin-Width

and formula for obtaining Number of Bins is given in equation (2).

$$No.OfBins = (max - min)/h \qquad (2)$$

where,
- max is the maximum value of the observation (here Comment Length)
- min is the minimum value of the observation (here Comment Length)

Here, the number of bins came out to be 659 approximately.

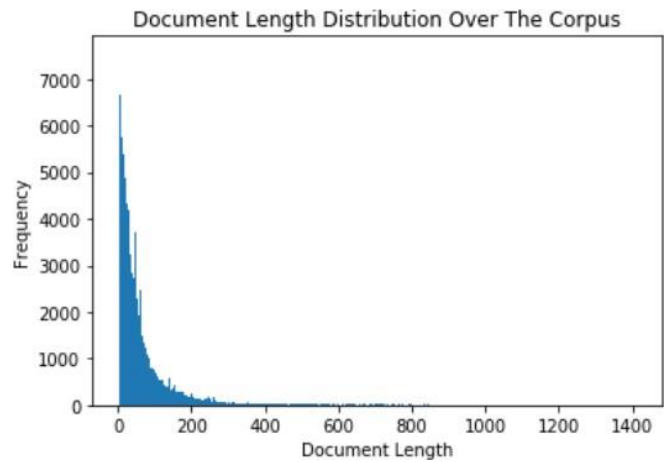

**Fig 2. Histogram showing Distribution of Comment Length over the whole corpus**

### 3.3    Text Preprocessing

The text preprocessing techniques followed before processing the text data are:

- **Removal of Punctuation**: All the punctuation marks in every comment are removed.
- **Lemmatisation**: Inflected forms of words which may be different verb forms or sigular/plural forms etc. are called lemma. For ex. go and gone are inflected forms or lemma of the word, gone. The process of grouping these lemma together is called Lemmatisation. So, Lemmatisation is performed for every comment.
- **Removal of Stopwords**: Frequently occurring common words like articles, prepositions etc. are called stopwords. So, stopwords are removed for each comment.

### 3.4    The Pipelines

6 pipelines are used where each pipeline corresponds to each label. With the help of these pipelines, 6 models are instantiated and trained separately.

- The 1st, 3rd and 5th Pipelines correspond to the labels toxic, obscene and insult respectively. The 3 stages of these pipelines are similar but these are trained separately. The stages of these pipelines are as follows:
    - **Bag-of-Words using Word Count Vectorizer**
      Bag-of-Words is a feature engineering technique in which a bag is maintained which contains all the different words present in the corpus. This bag is known as Vocabulary or Vocab. For each and every word present in the Vocabulary, counts of these words become the features for all the comments present in the corpus. A different and simpler example of Bag-of-Words is shown in Fig 3.

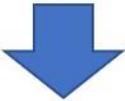

**Fig 3. Simpler Example of Bag-of-Words**

- **tf-idf Transformer**

The featured Bag-of-Words Model or Matrix for the whole corpus is transformed into a matrix whose every element is a product of Term Frequency (TF) and Inverse Document Frequency (IDF), combined together as tf-idf.

Term Frequency (TF) is defined as the ratio of the number of times a word or a term appears in a comment to the total number of words in the comment.

The formula for Inverse Document Frequency (IDF) is given in equation (3)

$$IDF = log(N/n) \qquad (3)$$

where,
- N is the total number of comments
- n is the number of comments a word has appeared in.

A different simpler example of tf-idf is shown in Fig 4.

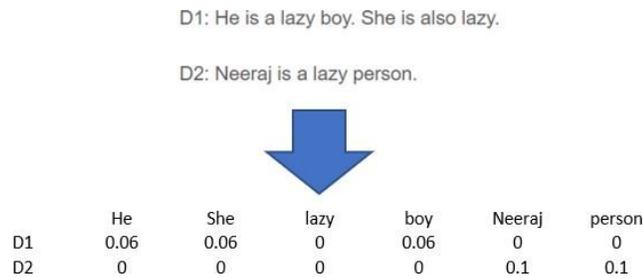

**Fig 4. Simpler Example of tf-idf**

- **Support Vector Machine Model with Linear Kernel**

After tf-idf transformation, a complete numeric featured dataset is obtained. Now, a Support Vector Machine Model is instantiated with 100 as the maximum number of iterations and C (penalty parameter) as 1.0.

Linear Support Vector Machine Algorithm:
1. The p-dimensional training instances (with p features) are assumed to be plotted in space.
2. A Hyperplane is predicted, which separates the different classes.
3. The best hyperplane should be selected finally, which maximizes the margin between data classes. The data points, influencing the hyperplane are known as Support Vectors.
4. The Large Margin Intuition for selection of best hyperplane for Linear SVM is given below:

$$\min_{\theta} C \sum_{i=1}^{m} \left[ y^{(i)} cost_1(\theta^T x^{(i)}) + (1 - y^{(i)}) cost_0(\theta^T x^{(i)}) \right] + \frac{1}{2} \sum_{i=1}^{n} \theta_j^2$$

where,
* $C$ is the penalty parameter
* theta is the parameter which needs to be optimized.

- The 2nd, 4th and 6th Pipelines correspond to the labels severe_toxic, threat and identity_hate respectively. Again the 3 components of these pipelines are similar but are trained separately. Also the 1st and 2nd stages of these pipelines are similar to those in the 1st, 3rd and 5th Pipelines. Only the 3rd stage is different and crucial.
  - **Bag-of-Words using Count Vectorizer**
  - **tf-idf Transformer**
  - **Decision Tree Classifter**
    
    After tf-idf transformation, a complete numeric featured dataset is obtained. Now, a Decision Tree Classifier is instantiated.
    
    Decision Tree Classifier Algorithm:
    1. The best feature of the dataset is placed at the root of the tree.
    2. The Training Samples are splitted into subsets such that each subset contains data with the same value for a feature.
    3. Steps 1 and and 2 are repeated on all the subsets until leaf nodes are found in all the branches of the tree.

## 4  Individual Training of the Pipelines

The dataset is 80-20 random splitted into Training and Testing (Validation) Sets. Out of 159571 instances, 127656 instances are used for training the 6 pipelines individually and the remaining 31915 instances are used for the individual and combined Validation and Performance Measure. So, the 6 pipelines are trained individually and tested.

## 5  Implementation Details

The text preprocessing, Bag-of-Words and tf-idf Transformer along with the training of the pipeline i.e., the training of the Machine Learning Models are implemented using Python's NLTK (Natural Language Toolkit) and are done on a machine with Intel(R) Core(TM) i5-8250U processor, CPU @ 1.60 GHz 1.80 GHz and 8 GB RAM, by Python's Scikit-Learn Machine Learning Toolbox.

## 6  Results

It consists of 2 subsections: Sub Section 6.1 explains the Individual Results and Collective Results are shown in Sub Section 6.2.

## 6.1 Individual Results

- Training Accuracy describes the accuracy achieved on the training set.
- Validation Accuracy describes the accuracy achieved on the Test Set.
- Precision is defined as the ratio of correctly predicted positive observations to the total predicted positive observations. The formula for Precision is shown in equation (4).

$$Precision = TP/TP + FP \qquad (4)$$

- The Sensitivity or Recall is defined as the proportion of correctly identified positives. The formula for Recall is given in equation (5).

$$Recall = TP/TP + FN \qquad (5)$$

- F1-Score is the Harmonic Mean of Precision and Recall.

After training, the pipelines are used for testing or validating the remaining 31915 samples. All the pipelines are allowed to give their predictions independently. But for the label severe_toxic, it is obvious that unless a comment is detected to be toxic, it has no chance of being severe_toxic. So based on the predictions made by the 1st Pipeline for the label toxic, those test instances which are not detected as toxic, are labelled 0, for the label severe toxic i.e., not detected as severe toxic. Hence, only for the label severe toxic, a 2nd check is done with reference to the prediction made by 1st Pipeline for the label toxic i.e., only those instances which are detected positive (Toxic) by 1st Pipeline are fed to the 2nd Pipeline for predictions. So, no Training Accuracy has been shown for the 2nd Pipeline for the label severe toxic.

The Training Accuracy, Validation Accuracy, Precision, Recall and F1-Score for all the pipelines/labels are tabulated in Table 1.

| Pipeline/Label | Training Accuracy | Validation Accuracy | Precision | Recall | F1-Score |
|---|---|---|---|---|---|
| 1st Pipeline/Toxic | 99.05% | 96.01% | 0.96 | 0.96 | 0.96 |
| 2nd Pipeline/Severe_Toxic | - | 98.85% | 0.99 | 0.99 | 0.99 |
| 3rd Pipeline/Obscene | 99.58% | 97.89% | 0.98 | 0.98 | 0.98 |
| 4th Pipeline/Threat | 99.99% | 99.66% | 1 | 1 | 1 |
| 5th Pipeline/Insult | 99.36% | 97.13% | 0.97 | 0.97 | 0.97 |
| 6th Pipeline/Identity_Hate | 99.99% | 98.96% | 0.99 | 0.99 | 0.99 |

**Table 1. Training Accuracy, Validation Accuracy, Precision, Recall and F1-Score for all the pipelines/labels**

- Area Under Receiver Operator Characteristic Curve (AUROC)
  ROC Curve is the plot of True Positive Rate vs False Positive Rate. The Area under ROC Curve should be greater than 0.5 to be termed as an Acceptable Test. Here, the ROC Curves are plotted by taking the predictions given by the pipelines on the Validation Set, rather than considering the distances of the predicted probabilities from the decision boundary.
  ROC Curves for all the pipelines/labels shown in Fig 5.

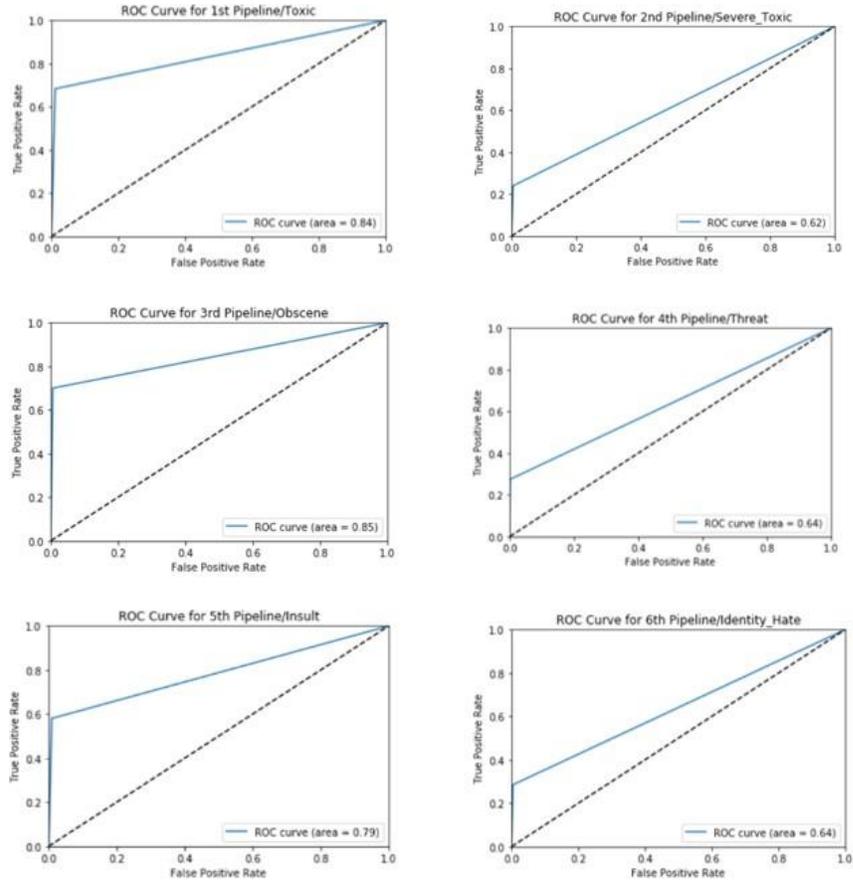

**Fig 5. ROC Curves and Area Under ROC Curves for all the pipelines/labels**

- Confusion Matrix Structure is shown in Table 2.

|  |  | Predicted Class |  |
|---|---|---|---|
|  |  | Class = Yes | Class = No |
| Actual Class | Class = Yes | True Positive | False Negative |
|  | Class = No | False Positive | True Negative |

**Table 2. Structure of Confusion Matrix**

Confusion Metrics for all the pipelines/labels are shown in Fig 6.

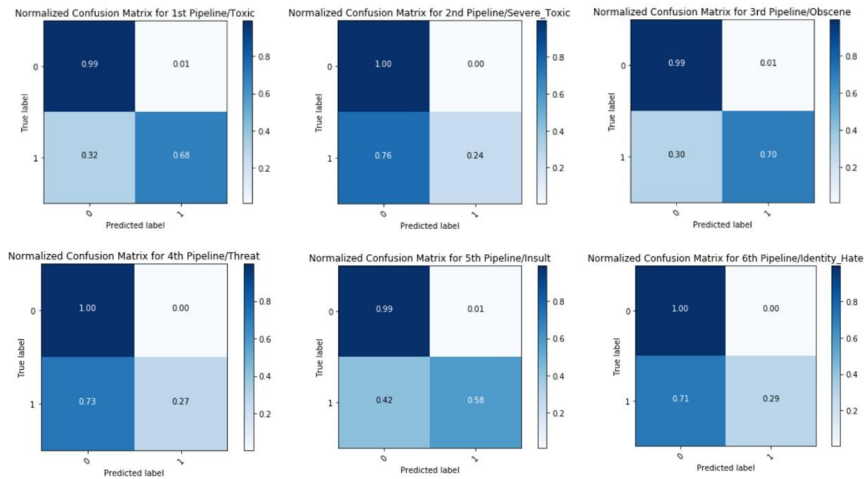

**Fig 6. Confusion Metrics for all the pipelines/labels**

### 6.2 Collective Results

- Mean Validation Accuracy is the average of the Validation Accuracies achieved by the 6 Pipeline Models. Hence, it is the Mean Validation Accuracy of the 6 Headed Model prepared.
  From this model, a Mean Validation Accuracy of 98.08% is achieved.
- Macro F1 is the average of the F1 Scores obtained by the 6 Pipeline Models. Hence, it is the Macro F1 of the 6 Headed Model prepared.
  From this model, a Macro F1 of 0.9817 is achieved.
- Macro AUC (Area Under ROC Curve) is the average of the AUCs achieved by the 6 Pipeline Models. Hence, it is the Macro AUC of the 6 Headed Model prepared.
  From this model, a Macro AUC of 0.73 is obtained.
- Absolute Validation Accuracy is the Validation Accuracy of the 6 Headed Model in which the accuracy is measured in such a way that, if all the predictions made by the 6 pipelines on the validation set i.e., the predictions for all the 6 labels, matches exactly with the true values of the 6 labels for any instance/sample, then the Model is said to be Absolutely Accurate.
  From this model, an Absolute Validation Accuracy of 91.61% is achieved.

A Direct Comparison of this model with the existing models has been done on different parameters and shown in Table 3.

| Comparison Parameters | Georgeakopoulos et al. [3] | Khieu et al. [4] | Chu et al. [5] | Kohli et al. [6] | Model |
|---|---|---|---|---|---|
| Methodology | Convolutional Neural Network | LSTM | CNN with character-level embeddings | LSTM with custom embeddings | tf-idf with 6 headed Machine Learning |
| Mean Validation Accuracy | 91.2% | 92.7% | 94% | 97.78% | 98.08% |
| Macro F1 | - | 0.706 | - | - | 0.9817 |
| Macro AUC | - | - | - | 0.7240 | 0.73 |

**Table 3. Comparison with Existing Models**

# 7 Conclusion

This paper proposed a Machine Learning Approach combined with Natural Language Processing for toxicity detection and its type identification in user comments. Finally, the Mean Validation Accuracy, so obtained, is 98.08% which is by far the highest ever numeric accuracy reached by any Comment Toxicity Detection Model. The research done in this paper is intended to enhance fair online talk and views sharing in social media.

A more robust model can be developed by applying Grid Search Algorithm on the same dataset over the Machine Learning Algorithms for every pipeline, being used in order to obtain more better results and accurate classifications.